\pdfoutput=1

\documentclass[11pt]{article}

\usepackage[final]{acl}

\usepackage{times}
\usepackage{latexsym}

\usepackage[T1]{fontenc}

\usepackage[utf8]{inputenc}

\usepackage{microtype}

\usepackage{inconsolata}

\usepackage{graphicx}
\usepackage{tikz}
\usepackage[edges]{forest}
\usepackage{tikzscale}
\usepackage{multirow}
\usepackage{colortbl}
\usepackage{xcolor}  
\usepackage[figuresleft]{rotating}
\usepackage{enumitem}
\usepackage{booktabs}
\usepackage{amssymb}
\usepackage{pifont}
\usepackage{comment}
\newcommand{\cmark}{\ding{51}}%
\newcommand{\xmark}{\ding{55}}%

\newcommand{\cut}[1]{}
\newcounter{tbsnr}
\newenvironment{tbs}
{\addtocounter{tbsnr}{1}\par\bigskip\noindent\fbox{\thetbsnr}
\hspace*{\fill}\begin{minipage}{7cm}\tt}
{\end{minipage}\hspace*{\fill}\bigskip}

\newcommand{\name}[0]{\textsc{Judge-Bench}}

\title{LLMs instead of Human Judges? \\A Large Scale Empirical Study across 20 NLP Evaluation Tasks}

\author{
 \textbf{Anna Bavaresco\textsuperscript{1}},
 \textbf{Raffaella Bernardi\textsuperscript{2}},
 \textbf{Leonardo Bertolazzi\textsuperscript{2}},
 \textbf{Desmond Elliott\textsuperscript{3}},
\\
 \textbf{Raquel Fern\'andez\textsuperscript{1}},
 \textbf{Albert Gatt\textsuperscript{4}},
 \textbf{Esam Ghaleb\textsuperscript{5}},
 \textbf{Mario Giulianelli\textsuperscript{6}},
\\
 \textbf{Michael Hanna\textsuperscript{1}},
 \textbf{Alexander Koller\textsuperscript{7}},
 \textbf{Andr\'e F. T. Martins\textsuperscript{8}},
 \textbf{Philipp Mondorf\textsuperscript{9}},
\\
 \textbf{Vera Neplenbroek\textsuperscript{1}},
 \textbf{Sandro Pezzelle\textsuperscript{1}},
 \textbf{Barbara Plank\textsuperscript{9}},
 \textbf{David Schlangen\textsuperscript{10}},
\\
 \textbf{Alessandro Suglia\textsuperscript{11}},
 \textbf{Aditya K Surikuchi\textsuperscript{1}},
 \textbf{Ece Takmaz\textsuperscript{4}},
 \textbf{Alberto Testoni\textsuperscript{12}}
\\
\\
 \textsuperscript{1}University of Amsterdam,
 \textsuperscript{2}University of Trento,
 \textsuperscript{3}University of Copenhagen, \\
 \textsuperscript{4}Utrecht University, 
 \textsuperscript{5}Max Planck Institute for Psycholinguistics,
  \textsuperscript{6}ETH Z\"urich,\\
 \textsuperscript{7}Saarland University, 
 \textsuperscript{8}Universidade de Lisboa \& Unbabel,
 \textsuperscript{9}LMU Munich \& MCML,\\
 \textsuperscript{10}University of Potsdam,
 \textsuperscript{11}Heriot-Watt University
 \textsuperscript{12}Amsterdam UMC\\
 \vspace*{5pt}
 \url{judgebench@googlegroups.com}\thanks{Authors listed in alphabetical order.}
}

\begin{document}
\maketitle
\begin{abstract}
There is an increasing trend towards evaluating NLP models with LLMs instead of human judgments, raising questions about the validity of these evaluations, as well as their reproducibility in the case of proprietary models.
We provide \name{}, an extensible collection of 20 NLP datasets with human annotations covering a broad range of evaluated properties and types of data, and comprehensively evaluate 11 current LLMs, covering both open-weight and proprietary models, for their ability to replicate the annotations.
Our evaluations show substantial variance across models and datasets. Models are reliable evaluators on some tasks, but overall display substantial variability depending on the property being evaluated, the expertise level of the human judges, and whether the language is human or model-generated. We conclude that LLMs should be carefully validated against human judgments before being used as evaluators.\looseness-1

\vspace{.11em}
\hspace{1.25em}\includegraphics[width=1.25em,height=1.25em]{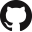}{\hspace{.75em}\parbox{\dimexpr\linewidth-2\fboxsep-2\fboxrule}
{\url{https://github.com/dmg-illc/JUDGE-BENCH}}}

\end{abstract}

\section{Introduction}

\begin{figure}[t]
    \centering
    \includegraphics[width=1\linewidth]{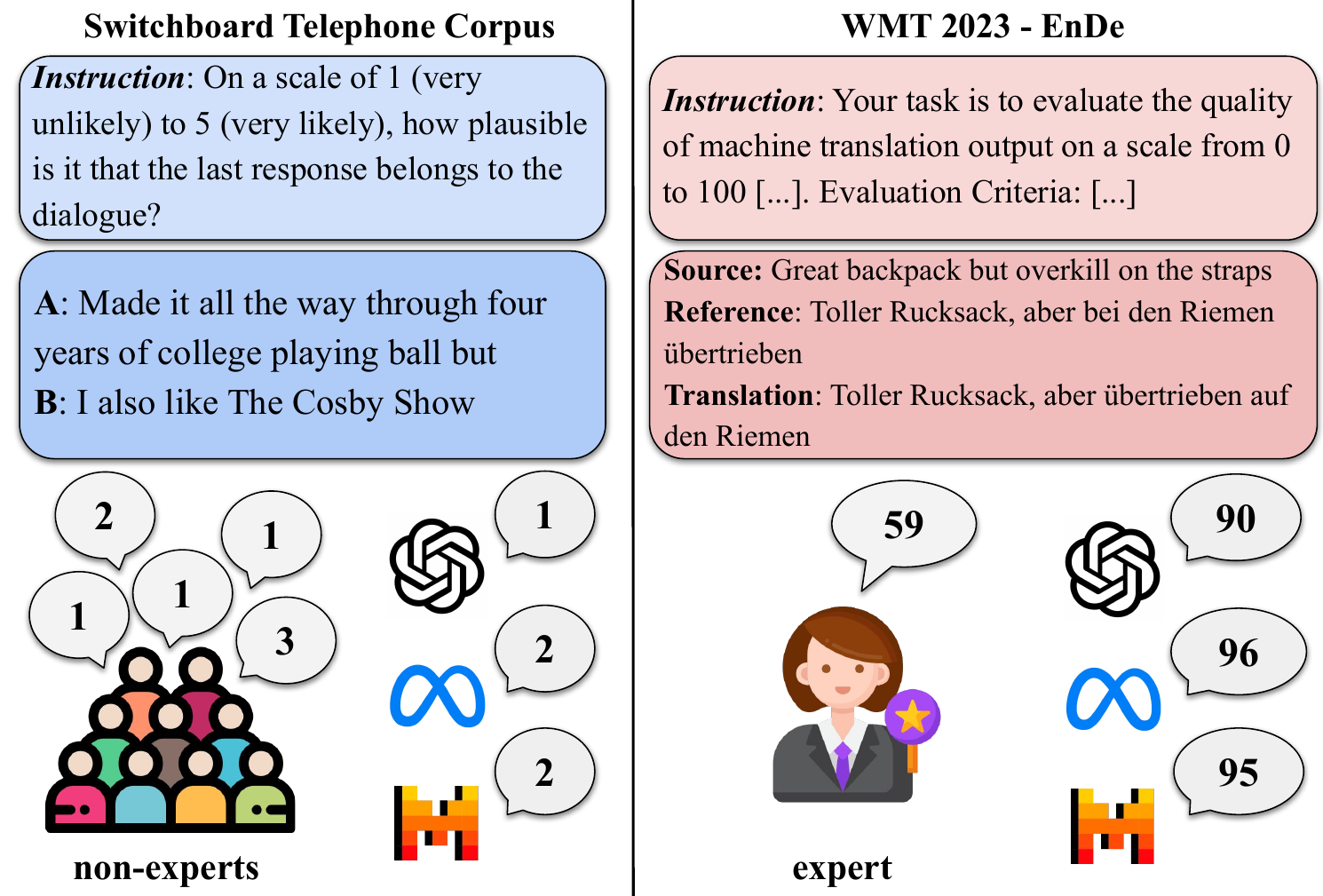}
    \caption{Evaluation by expert and non-expert human annotators and by LLMs for two tasks involving human-generated (left) and machine-generated text (right).
    }
    \label{fig:overview_intro}
    \vspace{-1em}
\end{figure}

For many natural language processing (NLP) tasks, the most informative evaluation is to ask humans to judge the model output. Such judgments are traditionally collected in lab experiments or through crowdsourcing, with either expert or non-expert annotators, as illustrated in Fig.~\ref{fig:overview_intro}. 
Recently, there has been a trend towards replacing human judgments with automatic assessments obtained via large language models \citep[LLMs;][\textit{inter alia}]{chiang-lee-2023-large,wang-etal-2023-chatgpt,liu-etal-2023-g,li-etal-2024-leveraging-large,zheng2024judging}.
For example, the LLM could be instructed to rate a response generated by a dialogue system for its perceived plausibility, on a scale from 1 to 5.
This drastically reduces the evaluation effort and is claimed to yield more reliable results across multiple evaluation rounds \cite{landwehr-etal-2023-memories, jiang-etal-2023-llm, reiter-blog24, dubois2024alpacafarm}.

At the same time, the use of LLMs as judges of linguistic output raises new concerns: LLMs may be prone to errors or systematic biases that differ from those of humans, especially on subtle tasks such as evaluating toxicity, or reasoning. 
This may distort evaluation results and lead to incorrect conclusions.
The problem is aggravated by explicit or implicit data leakage~\citep{balloccu-etal-2024-leak},  which undermines the ability to make broad, generalisable claims beyond the single specific dataset under analysis.
Specifically for closed models such as OpenAI's GPT series, there are serious reproducibility concerns, as LLMs may be retrained or retired at any time, making subsequent comparisons invalid or impossible.\looseness-1

Previous studies offer mixed evidence regarding the reliability of LLM evaluators.
Some research concludes that they are effective, correlating well with human judgments \citep{liu-etal-2023-g, zheng2024judging, chen-etal-2023-exploring-use,verga2024replacing,tornberg2023chatgpt, huang-etal-2024-chatgpt, naismith-etal-2023-automated, gilardi2023chatgpt,kocmi-federmann-2023-large}, albeit with some caveats \citep{wang-etal-2023-chatgpt, wu-aji-2025-style, hada-etal-2024-large, pavlovic-poesio-2024-effectiveness}. In some cases, LLM evaluators can also provide pairwise preference judgments \cite{kim-etal-2024-prometheus,liusie-etal-2024-llm,liu2024aligning,park2024paireval,tan2024judgebench}, or fine-grained evaluation beyond a single score, such as error spans \citep{fernandes2023devil,kocmi-federmann-2023-gemba}.
In contrast, some studies highlight substantial 
biases in LLMs' behaviour as evaluators, both as compared against human judgments \citep{koo-etal-2024-benchmarking, zeng2024evaluating, baris-schlicht-etal-2024-pitfalls} and through intrinsic analyses \cite{wang-etal-2024-large-language-models-fair,liu-etal-2024-llms-narcissistic,stureborg2024large}.
These discrepancies likely stem from the limitations of this previous work, which typically relies on a few datasets and models, often restricted to closed-source proprietary models. The observation of such limitations has motivated recent work to develop finetuning methods for LLM judges designed to overcome certain biases~\cite{zhu_judgelm_2025}.\looseness-1

In this paper, we examine how well current LLMs can approximate human evaluators on a large scale. We prompt 11 among the most recent open-weight and proprietary LLMs to generate judgments on 20 datasets with human annotations on a wide range of quality dimensions, prompt styles, and tasks.
Our evaluation goes beyond existing work by including a wide variety of \emph{datasets} that differ in the \emph{type of task} (e.g., translation, dialogue generation, etc.), the \emph{property} being judged (e.g., coherence, fluency, etc.), the \emph{type of judgments} (categorical or graded), and the \emph{expertise of human annotators} (experts or non-experts). 
We provide \name{}, a benchmark which includes, upon release, a total of over 70,000 test instances with associated human judgments with an extensible codebase.

Our results indicate that LLMs align well with human judgments on certain tasks, like instruction following.
However, their performance is \emph{inconsistent} across and within annotation tasks. Elicitation methods like Chain-of-Thought prompting \cite{wei2022} do not reliably improve agreement, in line with recent findings~\cite{sprague2024cot}. 
Some proprietary models---in particular, GPT-4o---align better to humans, but there is a rather small gap with large open-source models, holding promise for the reproducibility of future evaluation efforts.
Altogether, at the current stage of LLM development, we recommend validating LLM judges against task-specific human annotations before deploying them for any particular task. 

\section{Construction of \name}
\label{sec:datasets}

One key feature that differs across the datasets included in \name{} is the source of the data being evaluated, i.e., whether the items to be judged are generated by a model or produced by humans, as illustrated in Figure~\ref{fig:overview_intro}.

For model-generated items, the goal is to evaluate an NLP system. This includes both classic tasks such as machine translation or dialogue response generation, as well as less standard tasks for which automation has recently become an option thanks to LLMs, such as the generation of plans or logical arguments. 
For human-generated items, the goal is to assess properties of interest such as grammaticality or toxicity. 
This distinction allows us to understand whether LLMs have a positive bias towards machine-generated outputs---a tendency reported in prior work \cite{xu_pride_2024}.

The datasets we consider cover a wide span of properties of interest, ranging from grammaticality and toxicity to coherence, factual consistency, and verbosity, {\em inter alia}. Many properties are relevant across multiple tasks (e.g., fluency and coherence), while others are more task-specific (e.g., the success of a generated plan or the correctness of a multi-step mathematical reasoning trace).

\begin{table*}[ht!]
\resizebox{\textwidth}{!}{
\begin{tabular}{llllllll|cc}
\toprule
 & Dataset (\# properties judged) & GPT-4o & Llama-3.1-70B & Mixtral-8x22B & Gemini-1.5 & Mixtral-8x7B & Comm-R+ & $\sigma$ & \textit{UB} \\
\midrule
\multirow{18}{*}{\rotatebox[origin=c]{90}{Categorical Annotations}} 
 & \cellcolor{blue!25}CoLa (1) & 0.34  & 0.46  & 0.54  & 0.45  & \textbf{0.55}  & 0.12  & 0.16 & - \\
  & \cellcolor{blue!25}CoLa-grammar (63) & \textbf{0.47} ±0.22 & 0.28 ±0.24 & 0.28 ±0.23 & 0.26 ±0.24 & 0.21 ±0.18 & 0.13 ±0.14 & 0.14 & -  \\
 & \cellcolor{blue!25}ToxicChat (2) & \textbf{0.49} ±0.36 & 0.41 ±0.26 & 0.45 ±0.27 & 0.45 ±0.35 & 0.36 ±0.12 & 0.28 ±0.35 & 0.1 & - \\
 & \cellcolor{blue!25}LLMBar-natural (1) & \textbf{0.84}  & 0.8  & 0.72  & 0.79  & 0.54  & 0.56  & 0.13 & - \\
 & \cellcolor{blue!25}LLMBar-adversarial (1) & \textbf{0.58}  & 0.46  & 0.2  & 0.29  & 0.06  & 0.11  & 0.2 & - \\
  & \cellcolor{red!25}Persona Chat (2) & 0.24 ±0.34 & 0.24 ±0.33 & \textbf{0.58} ±0.59 & -0.03 ±0.04 & 0.54 ±0.65 & 0.48 ±0.74 & 0.2 & 0.88 \\
 & \cellcolor{red!25}Topical Chat (2) & \textbf{0.05} ±0.07 & -0.02 ±0.02 & -0.03 ±0.04 & -0.03 ±0.04 & 0.02 ±0.03 & 0.01 ±0.02 & 0.07 & 0.58 \\
 & \cellcolor{red!25}ROSCOE-GSM8K (2) & 0.59 ±0.35 & \textbf{0.64} ±0.27 & 0.62 ±0.38 & 0.6 ±0.24 & 0.58 ±0.36 & 0.0 & 0.15 & -  \\
 & \cellcolor{red!25}ROSCOE-eSNLI (2) & 0.29 ±0.06 & \textbf{0.38} ±0.08 & 0.13 ±0.13 & 0.11 ±0.18 & 0.1 ±0.11 & 0.03 ±0.05 & 0.14 & - \\
& \cellcolor{red!25}ROSCOE-DROP (2) & \textbf{0.29} ±0.08 & 0.27 ±0.07 & 0.2 ±0.12 & 0.08 ±0.05 & 0.13 ±0.21 & 0.03 ±0.04 & 0.13 & - \\
& \cellcolor{red!25}ROSCOE-CosmosQA (2) & 0.16 ±0.07 & \textbf{0.25} ±0.02 & 0.09 ±0.17 & 0.14 ±0.17 & 0.19 ±0.05 & -0.03 ±0.01 & 0.1 & - \\
 & \cellcolor{red!25}QAGS (1) & \textbf{0.72}  & 0.7  & 0.66  & 0.65  & 0.68  & 0.13  & 0.23 & 0.74 \\
  & \cellcolor{red!25}Medical-safety (2) & 0.01 ±0.03 & -0.03 ±0.06 & -0.02 ±0.09 & -0.03 ±0.08 & 0.0 ±0.06 & 0.01 ±0.02 & 0.03 & - \\
& \cellcolor{red!25}DICES-990 (1) & -0.24  & -0.17  & -0.16  & -0.12  & -0.2  & -0.09  & 0.05 & 0.27\\
 & \cellcolor{red!25}DICES-350-expert (1) & -0.2  & -0.13  & -0.15  & -0.03  & -0.11  & 0.01  & 0.08 & - \\
 & \cellcolor{red!25}DICES-350-crowdsourced (1) & -0.22  & -0.18  & -0.08  & -0.02  & -0.11  & -0.08  & 0.07 & 0.32\\
 & \cellcolor{red!25}Inferential strategies (1) & \textbf{0.42}  & 0.4  & 0.02  & 0.22  & 0.06  & -0.02  & 0.19 & 1.0\\
\midrule
& Average Cohen's $\kappa$ & 0.28 ±0.32 & 0.28 ±0.30 & 0.24 ±0.30&  0.22 ±0.28 & 0.21 ±0.28 & 0.10 ±0.18 & & \\
\midrule
\multirow{16}{*}{\rotatebox[origin=c]{90}{Graded Annotations}} 
 & \cellcolor{blue!25}Dailydialog (1) & \textbf{0.69}  & 0.6  & 0.55  & 0.63  & 0.63  & 0.52  & 0.06 & 0.79 \\
 & \cellcolor{blue!25}Switchboard (1) & \textbf{0.66}  & 0.45  & 0.63  & 0.59  & 0.56  & 0.36  & 0.11  & 0.8 \\
 & \cellcolor{red!25}Persona Chat (4) & \textbf{0.22} ±0.11 & -0.02 ±0.2 & 0.16 ±0.1 & 0.1 ±0.09 & 0.02 ±0.15 & 0.07 ±0.13 & 0.2 & 0.61 \\
 & \cellcolor{red!25}Topical Chat (4) & 0.26 ±0.03 & \textbf{0.28} ±0.1 & 0.13 ±0.04 & 0.17 ±0.12 & 0.21 ±0.18 & 0.14 ±0.05 & 0.07 & 0.56\\
 & \cellcolor{red!25}Recipe-generation (6) & \textbf{0.78} ±0.05 & 0.66 ±0.07 & 0.6 ±0.15 & 0.67 ±0.09 & 0.57 ±0.24 & 0.32 ±0.28 & 0.18  & 0.65 \\
  & \cellcolor{red!25}ROSCOE-GSM8K (2) & 0.82 ±0.12 & \textbf{0.83} ±0.11 & 0.81 ±0.14 & 0.81 ±0.12 & 0.79 ±0.13 & 0.68 ±0.2 & 0.15 & - \\
  & \cellcolor{red!25}ROSCOE-eSNLI (2) & \textbf{0.49} ±0.24 & 0.4 ±0.16 & 0.38 ±0.17 & 0.35 ±0.21 & 0.32 ±0.12 & 0.09 ±0.08 & 0.14 & - \\
 & \cellcolor{red!25}ROSCOE-DROP (2) & 0.57 ±0.22 & \textbf{0.59} ±0.16 & 0.44 ±0.15 & 0.44 ±0.13 & 0.32 ±0.12 & 0.21 ±0.22 & 0.13 & - \\
  & \cellcolor{red!25}ROSCOE-CosmosQA (2) & \textbf{0.57} ±0.18 & 0.55 ±0.18 & 0.51 ±0.16 & \textbf{0.57} ±0.17 & 0.53 ±0.21 & 0.33 ±0.25 & 0.1 & -  \\
 & \cellcolor{red!25}NewsRoom (4) & \textbf{0.59} ±0.02 & \textbf{0.59} ±0.03 & 0.44 ±0.05 & 0.55 ±0.03 & 0.5 ±0.07 & 0.36 ±0.06 & 0.1  & 0.62\\
 & \cellcolor{red!25}SummEval (4) & 0.35 ±0.06 & 0.44 ±0.14 & \textbf{0.54} ±0.08 & 0.38 ±0.02 & 0.48 ±0.02 & 0.19 ±0.06 & 0.13  & - \\
  & \cellcolor{red!25}WMT 2020 En-De (1) & \textbf{0.63}  & 0.37  & 0.51  & 0.46  & 0.2  & 0.42  & 0.15  & 0.81\\
 & \cellcolor{red!25}WMT 2020 Zh-En (1) & \textbf{0.54}  & 0.39  & 0.48  & 0.41  & 0.25  & 0.42  & 0.1  & 0.62\\
 & \cellcolor{red!25}WMT 2023 En-De (1) & 0.22  & 0.14  & \textbf{0.23}  & 0.16  & 0.17  & 0.22  & 0.04 & - \\
 & \cellcolor{red!25}WMT 2023 Zh-En (1) & 0.17  & 0.14  & \textbf{0.19}  & 0.14  & 0.15  & 0.15  & 0.02 & -  \\
\midrule
& Average Spearman's $\rho$ & 0.50 ±0.21 & 0.43 ±0.22 & 0.44 ±0.19&  0.43 ±0.21 & 0.38 ±0.22 & 0.30 ±0.17 & & \\
\bottomrule
\end{tabular}
}
\caption{Scores per dataset for the models with $\geq$98\% valid response rates (results for all models in Tab.~\ref{tab:all_models_results_table}, App.~\ref{appendix:additional_results}):
Cohen's kappa for categorical annotations and Spearman's correlation for graded annotations. Boldface marks best model performance per dataset. Spearman's correlations are generally significant ($p<0.05$), with the exception of the Persona Chat and Topical Chat datasets (see Tab.~\ref{tab:all_models_results_table} in Appendix~\ref{appendix:additional_results} for more details). Datasets with both categorical and graded annotations appear twice. Datasets in \colorbox{blue!25}{blue} concern human-generated language, while those in \colorbox{red!25}{red} concern model-generated text. `$\sigma$' denotes the standard deviation of the scores across models per dataset (averaged over properties if more than one is judged per dataset). Upper-bound estimates (\textit{UB}) indicate the agreement between individual and aggregated human judgments.}
\label{tab:results_table}
\vspace{-1em}
\end{table*}

Our study focuses on English datasets or language pairs which include English as one of the languages. We keep track of whether the original annotation guidelines are available and whether the annotations are provided by experts or non-experts. We retain all available individual annotations. 
Dataset information 
is summarised in Table~\ref{tab:dataset-info}, Appendix~\ref{sec:appendix-datasets}.
All 20 datasets are formatted following a precise data schema to facilitate the integration of additional datasets. This makes \name{} easily extensible. We provide more details about the data schema in Appendix~\ref{appendix:data-schema}.

\begin{figure}[h]
    \centering
    \includegraphics[width=0.99\linewidth]{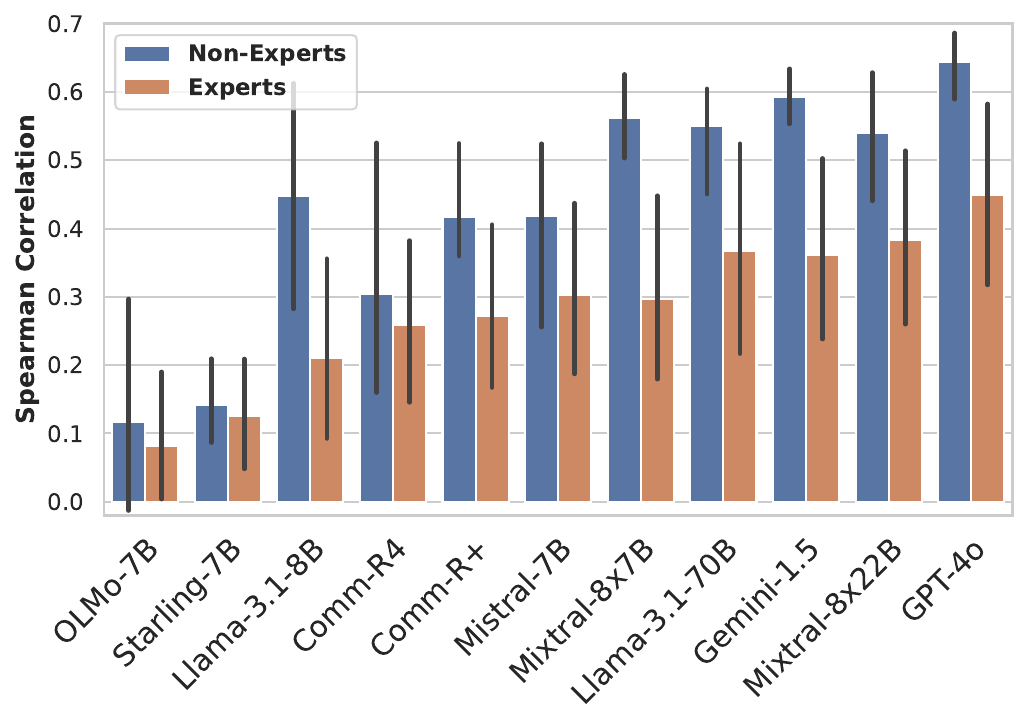}
    \caption{Average model correlation  
    with human experts vs.~non-experts in datasets with graded annotations.}
    \label{fig:experts_vs_crowdsource}
\end{figure}

\begin{figure}[h]
    \centering
    \includegraphics[width=0.99\linewidth]{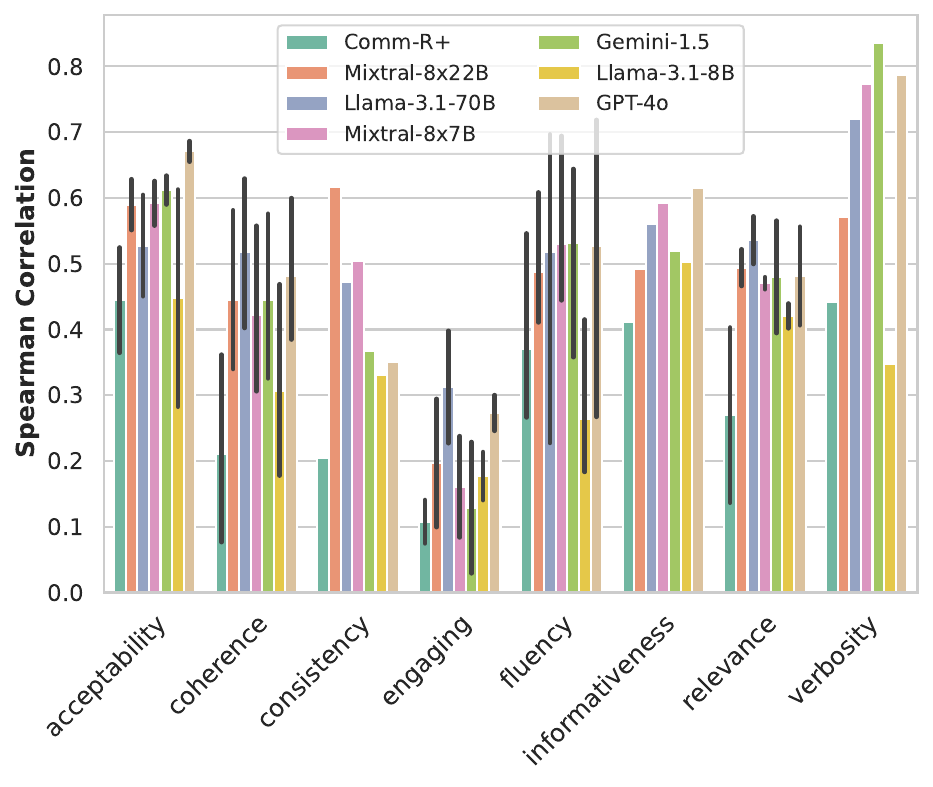}
    \vspace*{-.6cm}
    \caption{Correlation for properties with graded judgments. Averages and error bars when the property is present in more than one dataset.}
    \label{fig:across_categories}
\end{figure}

\section{Model Selection and Experiment Design}
\label{sec:models}

\paragraph{Models.}
We select representative proprietary and open-weight models of various sizes that show high performance across several tasks on the Open LLM and Chatbot Arena Leaderboards \cite{10.5555/3692070.3692401}: 
GPT-4o~\cite{gpt4o}, LLaMA-3.1 (8B and 70B; ~\citealt{llama3modelcard}), Gemini-1.5~\cite{reid2024gemini}, Mixtral (8x7B and 8x22B;~\citealt{jiang2024mixtral}), Command R
and Command R+~\cite{command_r_modelcard,command_rplus_modelcard}, OLMo~\cite{Groeneveld2023OLMo},
Starling-7B~\cite{starling2023}, and Mistral~\cite{jiang2023mistral}.
See Appendix~\ref{appendix:inference} for inference procedure details.

\paragraph{Prompts.}
Since most datasets include the original instructions used to gather human judgments, we use these instructions directly as prompts for the model, with additional guidelines to constrain the models' output and minimise verbosity: `\textit{Answer with one of \{\}. Do not explain your answer.}'
When the original instruction for collecting human judgments is unavailable, we create a prompt based on relevant information from the original paper, such as the task description and the definitions of the evaluation metrics.
We also experiment with alternative prompting strategies, including Chain-of-Thought, few-shot prompts, and prompt paraphrases. However, none of these strategies leads to systematic improvements. 
See Appendix~\ref{appendix:additional_results} for full details and results.
All prompts are provided in the codebase.

\paragraph{Evaluation.} 
Models do not always respond to the prompts as requested (e.g., they may refuse to answer if they perceive the prompt as sensitive). 
We therefore use the following evaluation protocol: 

\begin{itemize}[nosep,leftmargin=10pt]
    \item To obtain the same number of judgments across models for a given dataset, we replace invalid LLM responses with judgments randomly sampled from the relevant set of categorical or graded annotations. Figure~\ref{fig:appendix_models_response_rate} in Appendix~\ref{appendix:response_rates} shows the rate of valid responses per model. 
    \item Graded annotations, such as in WMT 2020~\citep{freitag-etal-2021-experts}, assess language quality on a continuous scale (e.g., a score from 0 to 100, or Likert-scale ratings), capturing varying degrees of fluency, adequacy, or overall translation quality; whereas categorical annotations, like those in CoLa~\citep{warstadt2019tacl}, involve binary judgments (e.g., grammatically acceptable or not). 
    For the former, we compute Spearman's correlation~($\rho$) between model and human judgments; 
    for the latter, we compute Cohen's $\kappa$.
    \item When multiple individual human judgments are available (typically three, see Table~\ref{tab:dataset-info} in Appendix~\ref{sec:appendix-datasets}), we estimate an upper bound by computing the average Spearman's $\rho$ or Cohen's $\kappa$ between bootstrapped single-rater responses and the aggregated responses across raters. Appendix~\ref{appendix:upper-bound} provides details on the upper bounds.
\end{itemize}

\section{Results}
\label{sec:results}
  
Scores vary substantially across models. For any given model, they vary both across datasets and properties being judged. 
Table~\ref{tab:results_table} presents detailed results for the 6 models that exhibit the largest rate of valid responses ($\geq$98\%).
GPT-4o ranks first across several evaluation scenarios, but the Llama-3.1-70B and Mixtral-8x22B open models are relatively close 
and outperform GPT-4o on some assessment types, such as categorical sentence acceptability (CoLa) and graded summary quality (SummEval).  
Overall, the high degree of variability is not fully accounted for by the inherent difficulty of the annotation tasks, as reflected in the human upper bound. Moreover, except for a few datasets (e.g., QAGS, Recipe-generation, and NewsRoom), model scores remain notably below the upper bound.

Among the property types with the lowest human-model alignment are toxicity and safety (in particular on DICES and Medical-safety), where model scores can be even negative and valid response rates particularly low (see Fig.~\ref{fig:response_rate_per_task} in Appendix~\ref{appendix:response_rates}). This is due in part to the guardrails associated with these tasks~\cite{weidinger2023sociotechnical}. 
We find that, especially in the medical domain, many models tend to provide explanations instead of producing a judgment (see Appendix~\ref{appendix:safety}). \looseness-1

Despite the high variability across models and datasets, we observe several notable trends. For graded annotations (Fig.~\ref{fig:experts_vs_crowdsource}), all models achieve higher correlations with annotations by non-expert human judges compared to expert annotators, echoing recent findings by \citet{aguda-etal-2024-large}. One possible explanation is that non-experts might rely on surface-level features, which could align more closely with the patterns LLMs are most attuned to, while experts apply stricter, domain-specific criteria. This remains speculative and calls for further investigation.
\looseness-1 

\begin{figure*}[th]
    \centering
    \includegraphics[width=1\linewidth]{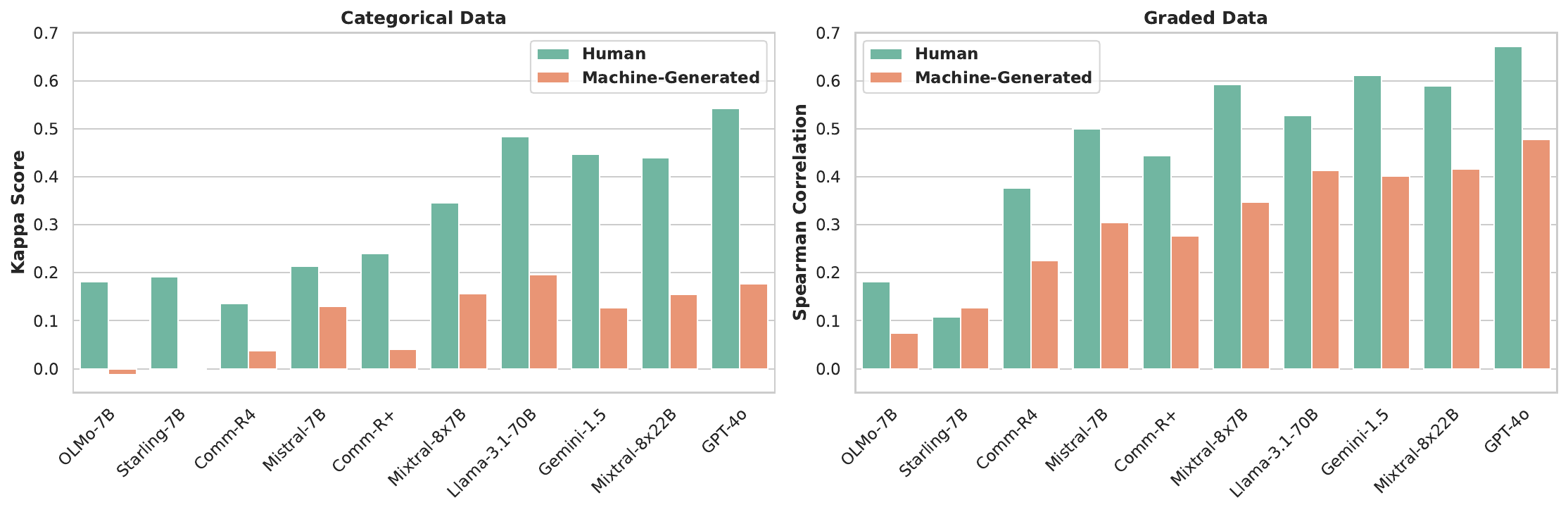}
    \caption{Scores (Cohen's $\kappa$ for categorical annotations and Spearman's correlation for graded annotations) on test items involving human language vs.~machine-generated outputs.}
    \label{fig:human_vs_machine_data}
\end{figure*}

Figure \ref{fig:across_categories} shows correlation results across different datasets for the subset of properties that exclusively have graded judgments. 
When applicable, we average results across datasets including annotations for the same property. We provide more details about these properties in Appendix~\ref{appendix:graded_judjs}. 
The proprietary models GPT-4o and Gemini-1.5 exhibit the highest scores when evaluating acceptability and verbosity, while the two Mixtral open models show the strongest correlations for coherence and consistency. Correlation with the engagingness property remains consistently low across all models.
Overall, \textit{no single model demonstrates a clear superiority} over others across all properties; instead, different quality dimensions are better assessed by different models. This calls into question the widespread practice of using a single model -- typically a proprietary one like those from the GPT family---to evaluate a diverse range of linguistic properties.

Finally, as shown in Figure~\ref{fig:human_vs_machine_data}, all models achieve better alignment with human judgments when evaluating human language than when assessing machine-generated text, both for categorical and graded annotations. This result aligns with the findings by \citet{xu_pride_2024}, suggesting that LLMs display a bias towards their own generation.
More broadly, this trend calls for caution when using LLMs to automatically evaluate the output of NLP systems.

\section{Conclusions}

In response to current trends in evaluation, in this paper we conducted a large-scale study of the correlation between human and LLM judgments across 20 datasets, 
considering factors such as the properties being assessed, the expertise level of the human judges, and whether the data is model- or human-generated. 
On some tasks, such as instruction following and the generation of mathematical reasoning traces, models can be reliably used as evaluators. Overall, however, models' agreement with human judgments varies widely across datasets, evaluated properties, and data sources; and depends on the level of expertise of human judges. Furthermore,  elicitation strategies such as Chain-of-Thought prompting do not consistently improve agreement levels, in line with recent findings~\cite{sprague2024cot}.
We recommend validation and calibration of LLMs against task-specific human judgments prior to their deployment as evaluators.
To facilitate this process, we release \name, a benchmark that enables systematic evaluation across a diverse range of tasks and is easily extensible to include any new task of interest.

\section*{Limitations}
One limitation of the experimental design of our work is that correlation with human judges may not be the most appropriate way to validate LLM evaluators. Indeed, there may be domains where human annotators and LLM evaluators appear aligned simply because they are affected by similar biases. Therefore, depending on the task at hand, it may be necessary to validate the reliability of human annotators as well.

Another limitation concerns the use of existing tasks and datasets without reassessing their quality or representativeness of actual downstream tasks.
While we did our best to select a wide set of tasks meaningful to the NLP community, we acknowledge that these tasks could not be equally meaningful for end-users, and that employing existing datasets could arguably lead to potential risks and shortcomings, such as data leakage.

In contrast to approaches that use LLMs for pairwise preference evaluation, e.g.,\ PairEval~\citep{park2024paireval} 
 or JudgeBench~\cite{tan2024judgebench},\footnote{Some time after an early version of this paper became available as a pre-print, accompanied by our Judge-Bench code, the independent work by~\citet{tan2024judgebench} appeared, which describes a benchmark that the authors named JudgeBench. This name clash is unfortunate, but since in the meantime our paper has seen some uptake, we have decided against trying to resolve it.} this paper focuses on evaluating the performance of LLMs on generating judgements for categorical and graded responses. 
 We leave the extension of \name{} to include pairwise preference evaluation and other recent evaluation methods, such as Prometheus 2~\citep{kim-etal-2024-prometheus}, for future work.

Finally, our work mostly focuses on English-language datasets---with the exception of datasets focussing specifically on machine-translation outputs. It remains to be seen whether LLMs' meta-evaluation abilities vary across different languages.

\section*{Acknowledgements}

This work emerged from discussions at a workshop organised by Raquel Fern\'andez and Sandro Pezzelle at MFO, the Oberwolfach Research Institute for Mathematics in the German Black Forest, on behalf of the ELLIS NLP programme. The event was funded by the state of Baden-W\"urttemberg (Germany) and organised in collaboration with the ELLIS Institute T\"ubingen and the Max Planck Institute for Intelligent Systems. We furthermore acknowledge our funders. In particular, 
AB, RF, and AT were supported by the European Research Council (ERC Consolidator Grant DREAM 819455 to RF), as well as BP (ERC Consolidator Grant DIALECT 101043235 to BP). DE was supported by a research grant (VIL53122) from VILLUM FONDEN. EG was supported by the Dutch Research Council (Gravitation grant 024.001.006 to the Language in Interaction consortium). MG was supported by an ETH Zurich Postdoctoral Fellowship. MH was supported in part by an OpenAI Superalignment Fellowship. AM was supported by the European Research Council (DECOLLAGE, ERC-2022-CoG 101088763) and by Funda\c{c}\~ao para a Ci\^encia e Tecnologia through contract UIDB/50008/2020.
We acknowledge ISCRA for awarding this project access to the LEONARDO supercomputer, owned by the EuroHPC Joint Undertaking, hosted by CINECA (Italy).

\bibliography{anthology,custom}

\appendix
\section*{Appendix}
\section{Datasets}
\label{sec:appendix-datasets}

This section provides brief descriptions of the datasets employed in our study. Table~\ref{tab:dataset-info} summarises relevant dataset information. Note that dataset sizes as reported in Table~\ref{tab:dataset-info} refer to the number of annotated samples (not to the total number of collected annotations) and might therefore differ from the figures reported in the original papers. 
Table~\ref{tab:krippendorf_table} reports Krippendorf's $\alpha$ for those datasets with multiple public human annotations. 

\paragraph{CoLa~\citep{warstadt2019tacl}.} The Corpus of Linguistic Acceptability (CoLA) consists of 10657 sentences from 23 linguistics publications, expertly annotated for acceptability (grammaticality) by their original authors.

\paragraph{CoLa-grammar~\citep{warstadt2020linguistic}.} The dataset consists of a grammatically annotated version of the CoLA development set. Each sentence in the CoLA development set is labelled with boolean features indicating the presence or absence of a particular grammatical construction (usually syntactic in nature). Two related sets of features are considered: 63 minor features correspond to fine-grained phenomena, and 15 major features correspond to broad classes of phenomena.    

\paragraph{ToxicChat~\cite{lin-etal-2023-toxicchat}.} collect binary judgments on the toxicity and `jailbreaking' nature (prompt hacks deliberately intended to bypass safety policies and induce models to generate unsafe content) of human prompts to LLMs. While the original dataset contains a mix of human- and automatically-annotated instances, here we only consider the human-annotated prompts. 

\begin{table*}[th]
\resizebox{\textwidth}{!}{
\begin{tabular}{lllrlcccc}\toprule
\textbf{Dataset}          & \textbf{Task} & \textbf{Size} & \textbf{\# Annot.} & \textbf{Type} & \textbf{Guidelines} & \textbf{Expert} & \textbf{Leaked}\\
\midrule
                              
CoLA~\cite{warstadt2019tacl}                              & Acceptability            & 1,043    & -     & Categorical     & \xmark    & \cmark     & \cmark        \\
CoLA-grammar~\cite{warstadt2020linguistic}                             & Acceptability            & 1,043  & -       & Categorical     & \xmark    & \cmark      & \cmark        \\
Switchboard~\cite{wallbridge22_interspeech}                        & Acceptability             & 100     & 3-6      & Graded                   & \cmark    & \xmark   &          \\
Dailydialog~\cite{wallbridge22_interspeech}                         & Acceptability              & 100     & 3-6      & Graded                   & \cmark      & \xmark     &        \\
Inferential strategies~\cite{mondorf-plank-2024-comparing}            & Reasoning              & 300   & -        & Categorical     & \cmark    & \cmark      &  \xmark\\
ROSCOE~\citep{golovneva2023roscoe}                     & Reasoning            & 756 & -        & Categorical + Graded     & \cmark  & \cmark     &        \\
Recipe-generation~\citep{stein-etal-2023-sentence}                 & Planning              & 52       & -     & Graded                    &     \cmark                       &        &     \\

Medical-safety~\citep{abercrombie-rieser-2022-risk}                  & Toxicity \& Safety           & 3,701   & -      & Preference               & \cmark      &  \cmark     &                            \\
DICES~\cite{NEURIPS2023_a74b697b}                  & Toxicity \& Safety           & 1,340   & \textasciitilde70 + \textasciitilde120      & Categorical              & \xmark    &  Mixed &                            \\
ToxicChat~\cite{lin-etal-2023-toxicchat}                        & Toxicity \& Safety                &  5,654 & -            & Categorical     & \xmark            &   \cmark &       \\
Topical Chat~\cite{mehri-eskenazi-2020-usr}                 & Dialogue                &    60    & 3       & Graded + Categorical       & \xmark   &    \cmark                      &         \\
Persona Chat~\cite{mehri-eskenazi-2020-usr}               & Dialogue                 &    60   & 3        &  Graded + Categorical      & \xmark    &   \cmark                        &        \\
WMT 2020 En-De~\citep{freitag-etal-2021-experts}               & Machine Translation           & 14,122  & 3      & Graded                   & \xmark     & \cmark        &     \\
WMT 2020 Zh-En~\citep{freitag-etal-2021-experts}                  & Machine Translation           & 19,974   & 3     & Graded                   & \xmark    &  \cmark      &       \\
WMT 2023 En-De~\citep{kocmi-etal-2023-findings}                 & Machine Translation           & 6,588   & -     & Graded                   & \xmark    &  \cmark    &      \\
WMT 2023 Zh-En~\citep{kocmi-etal-2023-findings}                & Machine Translation           & 13,245   & -     & Graded                   & \xmark    &    \cmark    &      \\
G-Eval / SummEval~\citep{liu-etal-2023-g}            & Summarisation            & 1,600    & -     & Graded                   & \cmark    &                            & \cmark      \\
QAGS~\citep{wang-etal-2020-asking}                        & Summarisation              & 953     & 3      & Categorical      & \cmark    & \xmark         &    \\
NewsRoom~\citep{grusky-etal-2018-newsroom}                   & Summarisation              & 420    & 3       & Graded                   & \cmark    & \xmark       & \cmark    \\
LLMBar~\citep{zeng2024evaluating}                   & Instruction Following              & 419   & -        & Categorical                   &  \cmark   &  \cmark      &   \xmark

\\\bottomrule

\end{tabular}
}
\caption{Overview of the main features of the datasets considered in the study. Note that `Size' refers to the number of annotated samples, not to the total number of human annotations. `\# Annot.' refers to the number of available individual annotations, if any, which we use to estimate the human upper bound. Note that datasets with only a single annotation per sample, or which only report the average over multiple annotations are not included in `\# Annot.'. Information on possible data leakage was retrieved from~\citet{balloccu-etal-2024-leak}.}
\label{tab:dataset-info}
\end{table*}

\paragraph{LLMBar~\citep{zeng2024evaluating}.} LLMBar is a dataset targeted at evaluating the instruction-following abilities of LLMs. Each entry of this dataset consists of an instruction paired with two different outputs, one correctly following the instruction and the other deviating from it. LLMBar has an \texttt{adversarial} split where deviating outputs are carefully constructed to `fool' LLM-based evaluators and a \texttt{natural} split where deviating outputs are more naturalistic.  

\paragraph{Topical Chat and Persona Chat~\citep{mehri-eskenazi-2020-usr}.} These datasets contain human judgments on the quality of machine- and human-generated responses based on the provided dialogue context. The annotated dialogues were selected from Topical Chat~\citep{gopalakrishnan_2019}---a dataset collecting human-human conversations on provided facts---and Persona Chat~\citep{zhang-etal-2018-personalizing}, which contains human-human persona-conditioned conversations. Each response is evaluated on 6 attributes: Understandable, Natural, Maintains Context, Interesting/Engaging, Uses Knowledge, and Overall Quality.

\paragraph{ROSCOE~\cite{golovneva2023roscoe}.} collect human judgments assessing the quality of GPT-3's reasonings. The output reasonings are elicited by inputting GPT-3 with questions selected from 4 commonly used reasoning datasets, i.e., CosmosQA~\citep{huang-etal-2019-cosmos}, DROP~\citep{dua-etal-2019-drop}, e-SNLI~\citep{camburu2018snli} and GSM8K~\citep{Cobbe2021TrainingVT}. While ROSCOE provides annotations on each step of the reasoning trace, here we only consider the global judgments over the whole reasoning. 

\paragraph{QAGS~\citep{wang-etal-2020-asking}.} QAGS consists of annotations judging the factual consistency of one-sentence model-generated summaries of news articles. The gold-standard summaries and articles are collected from CNN/DailyMail~\citep{hermann2015teaching} and XSUM~\citep{narayan-etal-2018-dont}. 

\paragraph{Medical-safety~\citep{abercrombie-rieser-2022-risk}.} This dataset consists of 3701 pairs of medical queries (collected from a subreddit on medical advice) and both machine-generated and human-generated answers. Queries were classified by human annotators according to their severity (from `Not medical' to `Serious', with `Serious' indicating that emergency care would be required) and answers were categorised based on their risk level (from `Non-medical' to `Diagnosis/Treatment').

\paragraph{DICES~\citep{NEURIPS2023_a74b697b}.} The DICES datasets consist of a series of machine-generated responses whose safety is judged based on the previous conversation turns (context). While the original dataset provides fine-grained annotations with answers to questions targeting specific aspects of safety, here we only consider the `overall' categorisation comprehensive of all aspects. In \texttt{DICES 990} safety is judged by crowdsourced annotators, whereas in \texttt{DICES 350} both expert and crowdsourced annotations are provided.  

\paragraph{Inferential strategies~\cite{mondorf-plank-2024-comparing}.} This dataset contains annotations on the logical validity of reasoning steps that models---in this case, \texttt{Llama-2-chat-hf3}~\citep{touvron2023}, \texttt{Mistral-7B-Instruct-v0.2}~\citep{jiang2023mistral} and \texttt{Zephyr-7b-beta}~\citep{tunstall2023zephyr}---generate when prompted to solve problems of propositional logic. Binary labels are assigned to each response, indicating whether the rationale provided by the model is sound (True) or not (False). Each model is assessed on 12 problems of propositional logic across 5 random seeds, resulting in a total of 60 responses per model.

\paragraph{Switchboard and Dailydialog~\citep{wallbridge22_interspeech}.} Switchboard includes acceptability judgments collected using stimuli from the Switchboard Telephone Corpus~\citep{godfrey1992switchboard}. More specifically, the judgments refer to how plausible it is that a specific response belongs to a telephonic dialogue. The same kind of judgments are provided for Dailydialog, which collects written dialogues intended to mimic conversations that could happen in real life. 

\paragraph{Recipe-generation~\citep{stein-etal-2023-sentence}.} This dataset contains human annotations assessing the quality of machine-generated recipes based on 6 attributes: grammar, fluency, verbosity, structure, success, overall.

\paragraph{NewsRoom~\citep{grusky-etal-2018-newsroom}.} This dataset includes human judgments on the quality of system-generated summaries of news articles. More specifically, annotators evaluated summaries across two semantic dimensions (informativeness and relevancy) and two syntactic dimensions (fluency and coherence).

\paragraph{SummEval and G-Eval~\citep{fabbri-summeval-2021,liu-etal-2023-g}.} These datasets include summaries generated by multiple recent summarisation models trained on the CNN/DailyMail dataset~\citep{hermann2015teaching}. Summaries are annotated by both expert judges and crowdsourced workers on 4 dimensions: coherence, consistency, fluency, relevance.

\paragraph{WMT 2020 En-De and Zh-En~\citep{freitag-etal-2021-experts}.} These datasets are a re-annotated version of the English-to-German and Chinese-to-English test sets taken from the WMT 2020 news translation task. The annotation was carried out by raters who are professional translators and native speakers of the target language using a Scalar Quality Metric (SQM) evaluation on a 0–6 rating scale.

\paragraph{WMT 2023 En-De and Zh-En~\citep{kocmi-etal-2023-findings}.} These datasets are the English-to-German and Chinese-to-English test sets taken from the General Machine Translation Task organised as part of the 2023 Conference on Machine Translation (WMT). In contrast to previous editions, the evaluation of translation quality was conducted by a professional or semi-professional annotator pool rather than utilising annotations from MTurk. Annotators were asked to provide a score between 0 and 100 on a sliding scale.

\begin{table}[ht!]
\resizebox{\linewidth}{!}{
\begin{tabular}{l|lc}
\toprule
 & Dataset & Krippendorf's $\alpha$\\
\midrule
\multirow{6}{*}{\rotatebox[origin=c]{90}{Categorical}} 
 & \cellcolor{red!25}Topical Chat & 0.08 \\
 & \cellcolor{red!25}QAGS & 0.49 \\
& \cellcolor{red!25}DICES-990 &  0.14 \\
 & \cellcolor{red!25}DICES-350-crowdsourced & 0.16\\
 & \cellcolor{red!25}Persona Chat & 0.33 \\
 & \cellcolor{red!25}Inferential strategies & 1.0\\
\midrule
\multirow{8}{*}{\rotatebox[origin=c]{90}{Graded}} 
 & \cellcolor{blue!25}Dailydialog & 0.59 \\
 & \cellcolor{blue!25}Switchboard & 0.57 \\
 & \cellcolor{red!25}Persona Chat & 0.33 \\
 & \cellcolor{red!25}Topical Chat & 0.08 \\
 & \cellcolor{red!25}Recipe-generation & 0.41 \\
 & \cellcolor{red!25}NewsRoom & 0.11  \\
 & \cellcolor{red!25}WMT 2020 En-De & 0.5 \\
 & \cellcolor{red!25}WMT 2020 Zh-En & 0.09 \\
\bottomrule
\end{tabular}
}
\caption{Inter-rater agreement for datasets with multiple human annotations. Datasets in \colorbox{blue!25}{blue} concern human-generated language, while those in \colorbox{red!25}{red} concern model-generated text. }
\label{tab:krippendorf_table}
\end{table}

\section{The \name{} Data Schema}
\label{appendix:data-schema}

To facilitate extending our benchmark, we adopt a shared schema used to pre-process all datasets. Our publicly available code base includes an example\footnote{\url{https://github.com/dmg-illc/JUDGE-BENCH/blob/master/data/example.json}} of this format as well as instructions on how to verify that newly added datasets comply with it.
    
The Json-based \name{} data schema ensures that the following fields are included for each dataset:
\begin{itemize}[nosep,itemsep=1pt,leftmargin=13pt]
        \item \texttt{dataset}: the name of the dataset;
        \item \texttt{dataset\_url}: the URL where the original dataset can be downloaded, as opensourced by their creators;
        \item \texttt{annotations}: an overview of the properties annotated for each dataset, along with information on how they are measured and prompt-like instructions similar to those originally given to the human annotators (when applicable);
        \item \texttt{instances}: dataset instances including the piece of text to be judged, aggregated human judgments and, when available, individual human annotations.
    \end{itemize}

\noindent
We note that, while we do not systematically explore inter-annotator variations at the instance level, the data schema we adopt allows for conducting this type of analysis in future work.

\section{Upper Bound Estimation for Model Correlations}
\label{appendix:upper-bound}

Whenever multiple human annotations were publicly available for a property (see Table~\ref{tab:krippendorf_table} for inter-annotator agreement scores), we computed upper-bound estimates for the correlations achievable by models. The intuition behind these estimates, borrowed from neuroscience \cite{nili2014toolbox}, is that the maximum correlation a model can achieve with aggregated human responses is bounded by the average correlation between single-participant responses and the aggregated responses across participants. 
We applied a similar logic to the human judgments used in the present study and combined it with a bootstrapping approach. For each annotated property, we bootstrapped single-participant responses by sampling $1000$ times from the available human responses, excluding data points where a single annotation was available. Next, we computed the alignment between each of the bootstrapped-participant arrays and the array of aggregated responses. Alignment was computed as Spearman's correlation for graded judgments and Cohen's kappa for categorical judgments. Finally, we estimated the upper bound as the average of the $1000$ alignment measures. In cases where alignment between bootstrapped and aggregated responses could not be computed---because the variance of the bootstrapped responses was null---values were replaced with an average of the `non-nan' correlations. 

We emphasise that these upper bounds are estimates and, as such, are subject to errors. Therefore, it may happen that model performance exceeds these upper bounds.

\section{Properties with Graded Judgments}
\label{appendix:graded_judjs}

In Figure~\ref{fig:across_categories}, we display results for a set of graded properties annotated in one or more of the datasets we consider. The properties are defined as follows:
\begin{itemize}[itemsep=1pt,leftmargin=10pt]
    \item \textit{Acceptability} refers to whether it is plausible or not that a response belongs to a telephonic dialogue and was annotated in Swithboard and Dailydialog;
    \item \textit{Coherence} was annotated for summaries and model-generated reasonings as part of the datasets NewsRoom, ROSCOE, and SummEval;
    \item \textit{Consistency} refers to the alignment between facts described in a summary and in its source text, and was annotated in SummEval; 
    \item \textit{Engaging} indicates whether a response generated in the context of a dialogue is dull or interesting and was annotated in TopicalChat and PersonaChat;
    \item \textit{Fluency} measures whether a piece of text is grammatically correct and well-formatted, and was annotated in NewsRoom, SummEval and Recipe-generation;
    \item \textit{Informativeness} refers to the extent to which a summary captures the key points of the full text, and was annotated for summaries as part of the NewsRoom dataset;
    \item \textit{Relevance} refers to whether a summary selects important information as opposed to including redundancies, and was annotated for NewsRoom and SummEval;
    \item \textit{Verbosity} indicates whether a generated recipe is concise and avoids unnecessary repetitions, and was annotated in Recipe-generation.
\end{itemize}

\section{Inference Details}
\label{appendix:inference}

All open-model checkpoints were obtained using the HuggingFace pipeline and we access all proprietary models using their corresponding API libraries. The proprietary models were accessed from 06-06-2024 to 13-06-2024, for standard prompting and from 09-10-2024 to 13-12-2024, for CoT prompting.
We obtain the model responses using greedy decoding, which we operationalise for the proprietary models by setting the temperature parameter to 0. We allow open models to generate a maximum of 25 new tokens and proprietary models to generate a maximum of 5 new tokens.
For CoT prompting, we allow for a maximum of 1000 new tokens.

We leverage Nvidia A100 (80 GB) GPUs for a total of $321$ compute hours. The cost of running experiments using Gemini-1.5-flash was €30.31, while the cost of experiments using GPT-4o was approximately \$565. 

\section{Valid Response Rates}
\label{appendix:response_rates}

\begin{figure}[ht]
    \centering
    \includegraphics[width=1\linewidth]{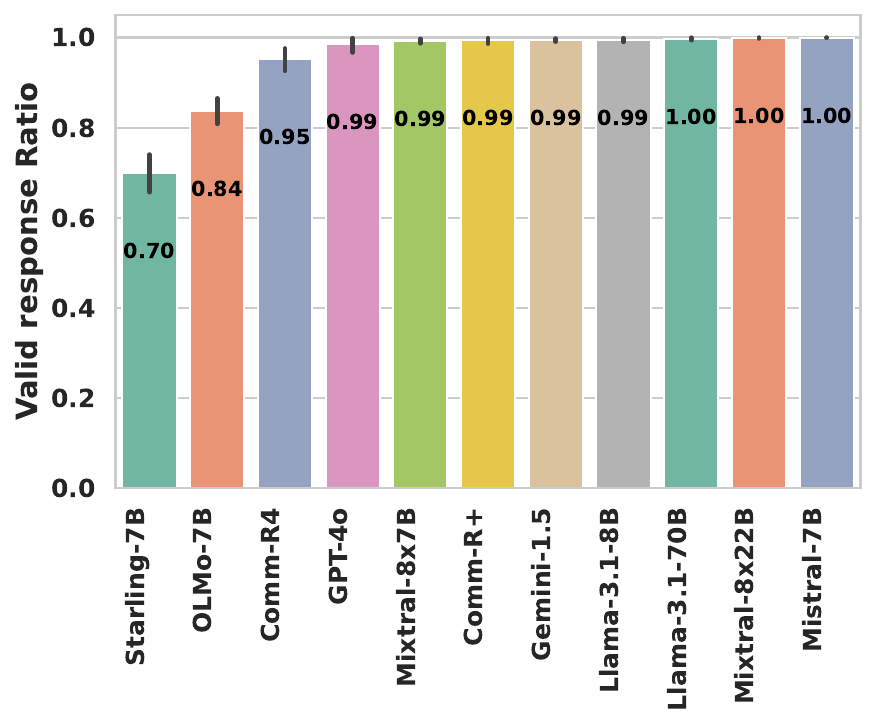}
    \caption{Valid response rate per model.}
    \label{fig:appendix_models_response_rate}
\end{figure}

\begin{figure*}[t]
    \centering
    \includegraphics[width=1\linewidth]{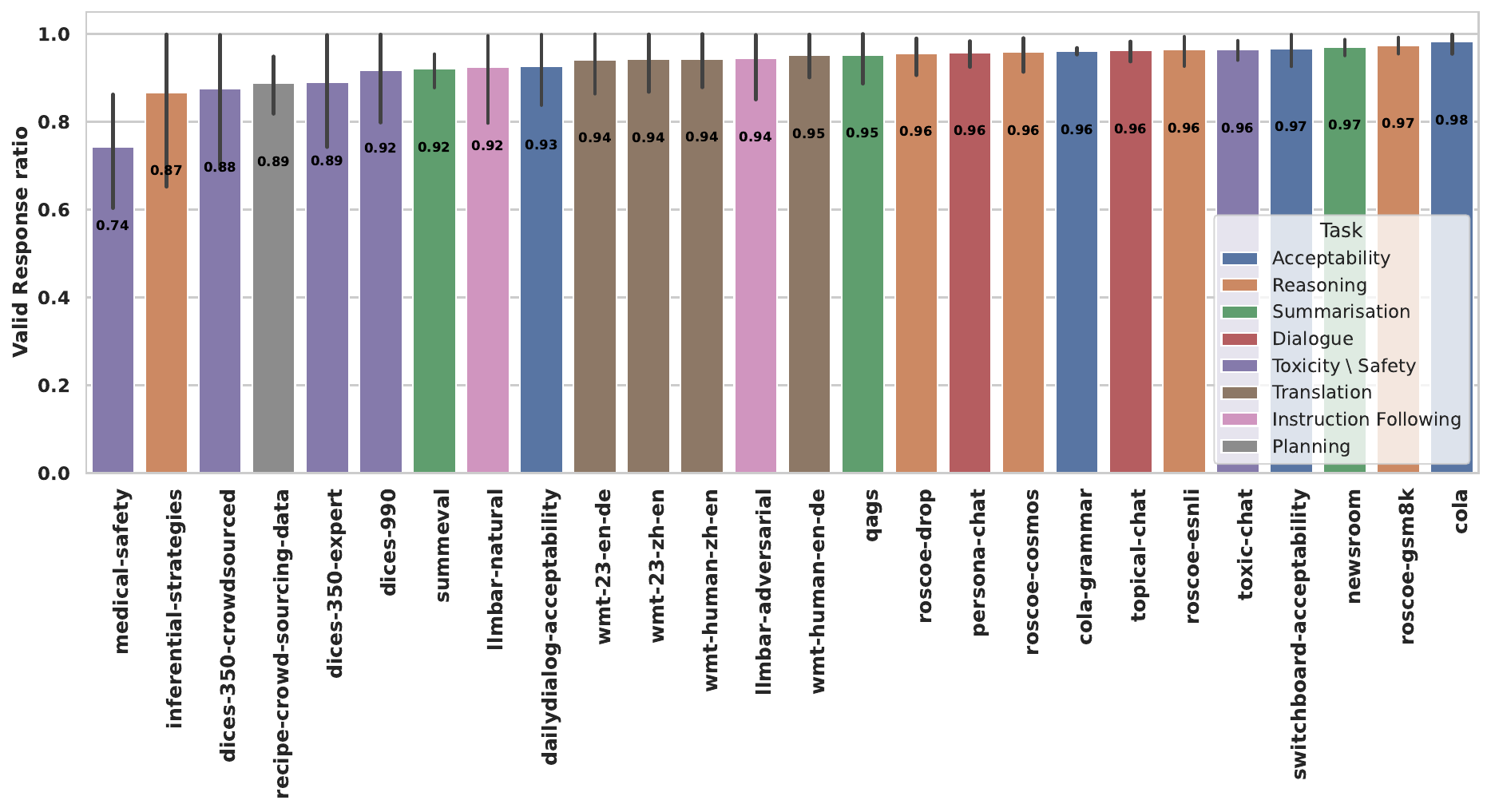}
    \caption{Average ratios of valid responses across datasets over the 11 models we tested.}
    \label{fig:response_rate_per_task}
\end{figure*}

Table~\ref{tab:all_models_resp_ratio}  reports the rate of valid responses for each model and dataset. Valid response rates are summarised per model and dataset in Figures~\ref{fig:appendix_models_response_rate} and \ref{fig:response_rate_per_task}.\looseness-1

\section{More Details on Toxicity and Safety Evaluation}
\label{appendix:safety}
For the Medical-safety dataset, models often refused to answer. Instead they tended to generate explanations, copy what they had in the prompt, or tried to be generally helpful because they saw that it was a medical issue. Since we take a random answer when no answer could be detected, this contributes to lower the results obtained on this task. Scores for the DICES dataset were also low, even though the valid response rate was high, because in this case there is the `Unsure' option, which (along with `Unsafe') models preferred over calling anything `Safe'. For ToxicChat, models performed reasonably well.

\section{Additional Results}
\label{appendix:additional_results}

In Table~\ref{tab:all_models_results_table} we report human-model alignment scores per dataset for all models tested, thus complementing Table~\ref{tab:results_table} in the paper.

\paragraph{Chain-of-Thought Prompts.}
For the results with CoT prompting, we use the same original instructions used to gather human judgments as prompts for the model but adapt the additional guidelines to emphasise multi-step reasoning rather than constrain the models' output. Specifically, we append the original instructions with the following additional guideline: `\textit{Always end your answer with either \{\} regarding the entire context. Let's think step by step.}', in which \textit{\{\}} is replaced with an enumeration of all possible answer labels formatted as `\textit{Therefore, \{label A\} is correct, or therefore, \{label B\} is correct, or therefore [...].}'. This also allows for automatically extracting the final answers from model responses during evaluation. 
In this study, we evaluate nine models and exclude Mixtral-8x22B and Comm-R+ due to computational constraints. For the CoLa-grammar dataset, we obtain GPT-4o responses only for ten percent of its instances (that are randomly sampled) to address the slow processing times and rate limitations. While CoT prompting leads to improved agreement scores and correlations when used with some models for certain datasets (see Table~\ref{tab:cot_models_results_table}), its overall effectiveness compared to the results obtained using standard prompts without CoT (see Table~\ref{tab:all_models_results_table}) is inconsistent.

\paragraph{Prompt Paraphrases.} We experiment with paraphrased prompts for three datasets that models struggle with: DICES-350-expert, WMT 2023 En-De, and WMT 2023 Zh-En. The paraphrase for dices-350-expert elaborates on the concept of safety, compared to its short original prompt, whereas the paraphrases for the WMT datasets are more concise regarding what comprises a good translation compared to the original.  We do not observe consistent improvements when using paraphrased prompts compared to the original prompts (Table~\ref{tab:alternative-prompting-strategies}).\looseness-1

\begin{table*}[th]
\centering \small
\begin{tabular}{cllll}
\toprule
& Prompt & Llama 3.1 8B & Llama 3.1 70B & Mixtral-8x7B \\ \midrule
\multirow{4}{*}{\textit{DICES-350-expert}} & Original         & 0.01         & -0.13         & -0.11        \\
& CoT              & -0.07        & -0.26         & -0.02        \\
& Few-shot         & 0.01         & -0.22         & -0.01        \\
& Paraphrase       & -0.13        & -0.36         & -0.09        \\\midrule
\multirow{4}{*}{\textit{WMT 2023 En-De}} & Original         & 0.08  (1)       & 0.14 (1)          & 0.17 (1)         \\
& CoT              & 0.18 (1)        & 0.16 (1)         & 0.20 (1)         \\
& Few-shot         & 0.19 (1)         & 0.20 (1)          & 0.20 (1)        \\
& Paraphrase       & 0.01 ±0.09 (3)   & 0.08 ± 0.12 (3)  & 0.14 ±0.05 (3)   \\\midrule
\multirow{4}{*}{\textit{WMT 2023 Zh-En}} & Original         & 0.02 (1)      & 0.14 (1)          & 0.15 (1)         \\
& CoT              & 0.13 (1)         & 0.13 (1)          & 0.16 (1)         \\
& Few-shot         & 0.15 (1)         & 0.14 (1)          & 0.16 (1)         \\
& Paraphrase       & 0.08 ±0.04 (3)  & 0.09 ±0.07 (2)    & 0.13 ±0.03 (3)  \\\bottomrule
\end{tabular}
\caption{Cohen's kappa for DICES-350-expert and Spearman’s correlation for two WMT 2023 datasets, comparing the original prompt and CoT prompt to few-shot prompts and prompt paraphrases for a selection of models. For datasets with more than one paraphrased prompt, we report the average and standard deviation across paraphrases. For Spearman's correlations, we report the number of significant correlations ($p<0.05$) for each model and dataset in brackets.}
\label{tab:alternative-prompting-strategies}
\end{table*}

\paragraph{Few-shot Prompts.} For the three datasets above---DICES-350-expert, WMT 2023 En-De, and WMT 2023 Zh-En---we also experiment with few-shot prompts (Table~\ref{tab:alternative-prompting-strategies}), where we provide the model with $6$ examples for DICES-350-expert, $3$ of safe conversations and $3$ of unsafe conversations, and $4$ examples for each WMT 2023 dataset, $2$ of high-scoring translations and $2$ of low-scoring translations. Using few-shot prompts does not improve correlations for dices-350-expert. On the WMT 2023  datasets, we observe higher correlations for Llama 3.1 8B but very moderate or no improvements on the other two models. Given that these improvements are inconsistent across datasets, we did not scale up the experiments to all $20$ datasets and $11$ models.

\begin{sidewaystable*}[ht!]\centering
\resizebox{0.9\textwidth}{!}{\begin{tabular}{lllllllllllll}
\toprule
Type & Dataset (\#Subtasks) & GPT-4o & Llama-3.1-70B & Mixtral-8x22B & Gemini-1.5 & Mixtral-8x7B & Comm-R+ & Comm-R4 & Llama-3.1-8B & Mistral-7B & Starling-7B & OLMo-7B \\
\midrule
\multirow{17}{*}{\rotatebox[origin=c]{90}{Categorical Annotations}} & \cellcolor{blue!25}CoLa (1) & 1.0 & 1.0 & 1.0 & 1.0 & 0.98 & 1.0 & 1.0 & 1.0 & 1.0 & 0.85 & 0.98 \\
  & \cellcolor{blue!25}CoLa-grammar (63) & 1.0 & 1.0 & 1.0 & 1.0 & 1.0 & 1.0 & 1.0±0.01 & 1.0 & 1.0 & 0.71±0.15 & 0.87±0.11 \\
 & \cellcolor{blue!25}LLMBar-natural (1) & 1.0 & 1.0 & 1.0 & 1.0 & 0.95 & 1.0 & 0.95 & 1.0 & 1.0 & 0.33 & 0.94 \\
 & \cellcolor{blue!25}LLMBar-adversarial (1) & 1.0 & 1.0 & 1.0 & 1.0 & 0.97 & 1.0 & 0.96 & 1.0 & 1.0 & 0.48 & 0.98 \\
 & \cellcolor{blue!25}ToxicChat (2) & 1.0 & 1.0 & 1.0 & 0.99 & 0.96±0.06 & 0.99 & 0.91±0.11 & 0.98 & 1.0 & 0.86±0.02 & 0.92±0.08 \\
  & \cellcolor{red!25}Persona Chat (2) & 1.0 & 1.0 & 1.0 & 1.0 & 0.98±0.02 & 1.0 & 0.89±0.15 & 1.0 & 1.0 & 0.96±0.01 & 0.58±0.24 \\
 & \cellcolor{red!25}Topical Chat (2) & 1.0 & 1.0 & 1.0 & 1.0 & 1.0 & 1.0 & 0.99±0.01 & 1.0 & 1.0 & 0.7±0.12 & 0.77±0.24 \\
 & \cellcolor{red!25}ROSCOE-GSM8K (2) & 1.0 & 1.0 & 1.0 & 1.0 & 1.0 & 1.0 & 1.0 & 1.0 & 1.0 & 0.92±0.01 & 0.8±0.23 \\
 & \cellcolor{red!25}ROSCOE-eSNLI (2) & 1.0 & 1.0 & 1.0 & 1.0 & 1.0 & 1.0 & 1.0 & 1.0 & 1.0 & 0.6±0.33 & 0.73±0.18 \\
 & \cellcolor{red!25}DICES-990 (1) & 1.0 & 1.0 & 1.0 & 0.99 & 0.98 & 1.0 & 1.0 & 1.0 & 1.0 & 0.77 & 0.35 \\
 & \cellcolor{red!25}Inferential strategies (1) & 1.0 & 1.0 & 1.0 & 1.0 & 0.99 & 0.97 & 1.0 & 1.0 & 1.0 & 0.05 & 0.53 \\
 & \cellcolor{red!25}ROSCOE-CosmosQA (2) & 1.0 & 1.0 & 1.0 & 1.0 & 1.0 & 1.0 & 1.0 & 1.0 & 1.0 & 0.49±0.45 & 0.75±0.19 \\
 & \cellcolor{red!25}QAGS (1) & 1.0 & 1.0 & 1.0 & 0.97 & 1.0 & 1.0 & 1.0 & 1.0 & 1.0 & 0.73 & 0.78 \\
 & \cellcolor{red!25}Medical-safety (2) & 0.35±0.37 & 0.96±0.02 & 0.97±0.04 & 0.97±0.04 & 0.85±0.1 & 0.78±0.31 & 0.33±0.47 & 0.89±0.11 & 1.0 & 0.22±0.08 & 0.85±0.19 \\
 & \cellcolor{red!25}DICES-350-expert (1) & 1.0 & 1.0 & 1.0 & 0.99 & 1.0 & 0.98 & 0.99 & 1.0 & 1.0 & 0.55 & 0.27 \\
 & \cellcolor{red!25}DICES-350-crowdsourced (1) & 1.0 & 1.0 & 1.0 & 0.99 & 0.99 & 0.98 & 1.0 & 1.0 & 1.0 & 0.51 & 0.16 \\
 & \cellcolor{red!25}ROSCOE-DROP (2) & 1.0 & 1.0 & 1.0 & 1.0 & 1.0 & 1.0 & 1.0 & 1.0 & 1.0 & 0.51±0.51 & 0.68±0.26 \\\midrule
 \multirow{13}{*}{\rotatebox[origin=c]{90}{Graded Annotations}}
 & \cellcolor{blue!25}Dailydialog (1) & 1.0 & 1.0 & 1.0 & 0.99 & 1.0 & 1.0 & 0.69 & 1.0 & 1.0 & 0.89 & 0.62 \\
 & \cellcolor{blue!25}Switchboard (1) & 1.0 & 1.0 & 1.0 & 1.0 & 0.99 & 1.0 & 0.93 & 1.0 & 1.0 & 0.95 & 0.77 \\
 & \cellcolor{red!25}Persona Chat (4) & 1.0 & 1.0 & 1.0 & 1.0 & 1.0 & 1.0 & 0.97±0.03 & 1.0 & 1.0 & 0.71±0.27 & 0.92±0.15 \\
 & \cellcolor{red!25}Topical Chat (4) & 1.0 & 1.0 & 1.0 & 1.0 & 1.0 & 1.0 & 0.99±0.01 & 1.0 & 1.0 & 0.75±0.1 & 0.91±0.07 \\
 & \cellcolor{red!25}Recipe-generation (6) & 1.0 & 1.0 & 1.0 & 1.0 & 1.0±0.01 & 1.0 & 0.67±0.2 & 1.0 & 1.0 & 0.11±0.16 & 0.98±0.01 \\
 & \cellcolor{red!25}ROSCOE-CosmosQA (2) & 1.0 & 1.0 & 1.0 & 1.0 & 1.0 & 1.0 & 0.99±0.01 & 1.0 & 1.0 & 0.97±0.01 & 0.89 \\
 & \cellcolor{red!25}ROSCOE-DROP (2) & 1.0 & 1.0 & 1.0 & 1.0 & 1.0 & 1.0 & 1.0 & 1.0 & 1.0 & 0.99±0.01 & 0.85±0.11 \\
 & \cellcolor{red!25}ROSCOE-eSNLI (2) & 1.0 & 1.0 & 1.0 & 1.0 & 1.0 & 1.0 & 1.0 & 1.0 & 1.0 & 1.0 & 0.89 \\
 & \cellcolor{red!25}ROSCOE-GSM8K (2) & 1.0 & 1.0 & 1.0 & 1.0 & 1.0 & 1.0 & 0.98 & 1.0 & 1.0 & 0.84±0.02 & 0.91±0.06 \\
 & \cellcolor{red!25}NewsRoom (4) & 1.0 & 1.0 & 0.98±0.01 & 0.99 & 1.0 & 1.0 & 1.0 & 1.0 & 1.0 & 0.89±0.1 & 0.83±0.04 \\
 & \cellcolor{red!25}SummEval (4) & 0.87±0.13 & 0.94±0.06 & 1.0 & 0.9±0.06 & 1.0 & 1.0 & 0.72±0.3 & 0.94±0.08 & 1.0 & 0.96±0.04 & 0.79±0.05 \\
  & \cellcolor{red!25}WMT 2020 En-De (1) & 1.0 & 1.0 & 1.0 & 0.99 & 0.87 & 1.0 & 1.0 & 1.0 & 1.0 & 0.85 & 0.75 \\
 & \cellcolor{red!25}WMT 2020 Zh-En (1) & 1.0 & 1.0 & 1.0 & 1.0 & 0.87 & 1.0 & 1.0 & 1.0 & 1.0 & 0.81 & 0.7 \\
 & \cellcolor{red!25}WMT 2023 En-De (1) & 1.0 & 1.0 & 1.0 & 1.0 & 1.0 & 1.0 & 1.0 & 1.0 & 1.0 & 0.79 & 0.58 \\
 & \cellcolor{red!25}WMT 2023 Zh-En (1) & 1.0 & 1.0 & 1.0 & 1.0 & 0.99 & 1.0 & 1.0 & 1.0 & 1.0 & 0.78 & 0.61 \\
\bottomrule
\end{tabular}}
\caption{\label{tab:all_models_resp_ratio} Ratios of valid responses per dataset for all models we evaluate.}
\end{sidewaystable*}

\begin{sidewaystable*}[ht!]\centering
\resizebox{0.9\textwidth}{!}{\begin{tabular}{lllllllllllll}
\toprule
Type & Dataset (\# properties judged) & GPT-4o & Llama-3.1-70B & Mixtral-8x22B & Gemini-1.5 & Mixtral-8x7B & Comm-R+ & Comm-R4 & Llama-3.1-8B & Mistral-7B & Starling-7B & OLMo-7B \\
\midrule
\multirow{17}{*}{\rotatebox[origin=c]{90}{Categorical Annotations}}
 & \cellcolor{blue!25}CoLa (1) & 0.34  & 0.46  & 0.54  & 0.45  & 0.55  & 0.12  & 0.01  & 0.42  & 0.43  & 0.45  & 0.42  \\
 & \cellcolor{blue!25}CoLa-grammar (63) & 0.47 ±0.22 & 0.28 ±0.24 & 0.28 ±0.23 & 0.26 ±0.24 & 0.21 ±0.18 & 0.13 ±0.14 & 0.08 ±0.1 & 0.1 ±0.14 & 0.09 ±0.13 & 0.07 ±0.08 & 0.04 ±0.06 \\
 & \cellcolor{blue!25}LLMBar-natural (1) & 0.84  & 0.8  & 0.72  & 0.79  & 0.54  & 0.56  & 0.59  & 0.57  & 0.3  & 0.28  & 0.24  \\
 & \cellcolor{blue!25}LLMBar-adversarial (1) & 0.58  & 0.46  & 0.2  & 0.29  & 0.06  & 0.11  & -0.2  & -0.18  & -0.2  & -0.12  & -0.1  \\
 & \cellcolor{blue!25}ToxicChat (2) & 0.49 ±0.36 & 0.41 ±0.26 & 0.45 ±0.27 & 0.45 ±0.35 & 0.36 ±0.12 & 0.28 ±0.35 & 0.2 ±0.21 & 0.34 ±0.29 & 0.45 ±0.18 & 0.27 ±0.26 & 0.3 ±0.13 \\
 & \cellcolor{red!25}Persona Chat (2) & 0.24 ±0.34 & 0.24 ±0.33 & 0.58 ±0.59 & -0.03 ±0.04 & 0.54 ±0.65 & 0.48 ±0.74 & 0.01 ±0.01 & 0.5 ±0.7 & 0.47 ±0.75 & -0.03 ±0.04 & 0.02 ±0.03 \\
 & \cellcolor{red!25}Topical Chat (2) & 0.05 ±0.07 & -0.02 ±0.02 & -0.03 ±0.04 & -0.03 ±0.04 & 0.02 ±0.03 & 0.01 ±0.02 & 0.01 ±0.01 & 0.57 ±0.61 & -0.03 ±0.05 & 0.04 ±0.06 & 0.03 ±0.04 \\
 & \cellcolor{red!25}ROSCOE-GSM8K (2) & 0.59 ±0.35 & 0.64 ±0.27 & 0.62 ±0.38 & 0.6 ±0.24 & 0.58 ±0.36 & 0.0 & 0.21 ±0.03 & 0.36 ±0.31 & 0.47 ±0.34 & -0.03 ±0.01 & -0.01 ±0.02 \\
 & \cellcolor{red!25}ROSCOE-eSNLI (2) & 0.29 ±0.06 & 0.38 ±0.08 & 0.13 ±0.13 & 0.11 ±0.18 & 0.1 ±0.11 & 0.03 ±0.05 & -0.01 ±0.01 & 0.14 ±0.2 & 0.02 ±0.09 & 0.01 ±0.07 & -0.04 ±0.09 \\
 & \cellcolor{red!25}ROSCOE-DROP (2) & 0.29 ±0.08 & 0.27 ±0.07 & 0.2 ±0.12 & 0.08 ±0.05 & 0.13 ±0.21 & 0.03 ±0.04 & 0.02 ±0.07 & 0.02 ±0.02 & 0.09 ±0.08 & 0.01 ±0.03 & 0.0 ±0.01 \\
 & \cellcolor{red!25}ROSCOE-CosmosQA (2) & 0.16 ±0.07 & 0.25 ±0.02 & 0.09 ±0.17 & 0.14 ±0.17 & 0.19 ±0.05 & -0.03 ±0.01 & -0.01 ±0.02 & 0.08 ±0.11 & 0.29 ±0.03 & 0.03 & -0.18 \\
 & \cellcolor{red!25}QAGS (1) & 0.72  & 0.7  & 0.66  & 0.65  & 0.68  & 0.13  & 0.33  & 0.58  & 0.43  & 0.02  & 0.11  \\
 & \cellcolor{red!25}Medical-safety (2) & 0.01 ±0.03 & -0.03 ±0.06 & -0.02 ±0.09 & -0.03 ±0.08 & 0.0 ±0.06 & 0.01 ±0.02 & 0.01 ±0.01 & 0.01 & -0.03 ±0.12 & 0.0 ±0.02 & -0.02 ±0.07 \\ 
 & \cellcolor{red!25}DICES-990 (1) & -0.24  & -0.17  & -0.16  & -0.12  & -0.2  & -0.09  & -0.02  & -0.11  & -0.12  & -0.05  & 0.0  \\
 & \cellcolor{red!25}DICES-350-expert (1) & -0.2  & -0.13  & -0.15  & -0.03  & -0.11  & 0.01  & 0.01  & 0.01  & 0.01  & 0.01  & -0.06  \\
 & \cellcolor{red!25}DICES-350-crowdsourced (1) & -0.22  & -0.18  & -0.08  & -0.02  & -0.11  & -0.08  & 0.01  & -0.05  & -0.04  & 0.01  & -0.03  \\
 & \cellcolor{red!25}Inferential strategies (1) & 0.42  & 0.4  & 0.02  & 0.22  & 0.06  & -0.02  & -0.12  & 0.13  & 0.01  & 0.01  & 0.04  \\\midrule
 \multirow{13}{*}{\rotatebox[origin=c]{90}{Graded Annotations}}
 & \cellcolor{blue!25}Dailydialog (1) & 0.69 (1)  & 0.6 (1)  & 0.55 (1)  & 0.63 (1)  & 0.63 (1)  & 0.52 (1)  & 0.23 (1)  & 0.61 (1)  & 0.48 (1)  & 0.09 (0)  & 0.07 (0)  \\
 & \cellcolor{blue!25}Switchboard (1) & 0.66 (1)  & 0.45 (1)  & 0.63 (1)  & 0.59 (1)  & 0.56 (1)  & 0.36 (1)  & 0.53 (1)  & 0.28 (1)  & 0.52 (1)  & 0.13 (0)  & 0.3 (1)  \\
 & \cellcolor{red!25}Persona Chat (4) & 0.22 ±0.11 (2) & -0.02 ±0.2 (0) & 0.16 ±0.1 (1) & 0.1 ±0.09 (0) & 0.02 ±0.15 (0) & 0.07 ±0.13 (0) & 0.05 ±0.2 (0) & -0.02 ±0.14 (0) & -0.09 ±0.17 (1) & 0.03 ±0.13 (0) & -0.06 ±0.14 (0) \\
 & \cellcolor{red!25}Topical Chat (4) & 0.26 ±0.03 (2) & 0.28 ±0.1 (2) & 0.13 ±0.04 (0) & 0.17 ±0.12 (1) & 0.21 ±0.18 (1) & 0.14 ±0.05 (0) & 0.07 ±0.07 (0) & 0.15 ±0.13 (0) & 0.29 ±0.11 (3) & 0.14 ±0.16 (1) & 0.08 ±0.21 (1) \\
 & \cellcolor{red!25}Recipe-generation (6) & 0.78 ±0.05 (6) & 0.66 ±0.07 (6) & 0.6 ±0.15 (6) & 0.67 ±0.09 (5) & 0.57 ±0.24 (5) & 0.32 ±0.28 (5) & 0.06 ±0.26 (3) & 0.34 ±0.09 (5) & 0.28 ±0.08 (4) & 0.04 ±0.17 (1) & 0.1 ±0.08 (0) \\
 & \cellcolor{red!25}ROSCOE-GSM8K (2) & 0.82 ±0.12 (2) & 0.83 ±0.11 (2) & 0.81 ±0.14 (2) & 0.81 ±0.12 (2) & 0.79 ±0.13 (2) & 0.68 ±0.2 (2) & 0.7 ±0.08 (2) & 0.76 ±0.15 (2) & 0.63 ±0.18 (2) & 0.46 ±0.13 (2) & 0.1 ±0.07 (1) \\
 & \cellcolor{red!25}ROSCOE-eSNLI (2) & 0.49 ±0.24 (2) & 0.4 ±0.16 (2) & 0.38 ±0.17 (2) & 0.35 ±0.21 (2) & 0.32 ±0.12 (2) & 0.09 ±0.08 (0) & 0.28 ±0.21 (1) & 0.19 ±0.16 (1) & 0.32 ±0.12 (2) & 0.11 ±0.06 (0) & 0.11 ±0.17 (1) \\
 & \cellcolor{red!25}ROSCOE-DROP (2) & 0.57 ±0.22 (2) & 0.59 ±0.16 (2) & 0.44 ±0.15 (2) & 0.44 ±0.13 (2) & 0.32 ±0.12 (2) & 0.21 ±0.22 (1) & 0.37 ±0.18 (2) & 0.23 ±0.1 (2) & 0.22 ±0.22 (1) & 0.16 ±0.17 (1) & 0.15 ±0.21 (1) \\
 & \cellcolor{red!25}ROSCOE-CosmosQA (2) & 0.57 ±0.18 (2) & 0.55 ±0.18 (2) & 0.51 ±0.16 (2) & 0.57 ±0.17 (2) & 0.53 ±0.21 (2) & 0.33 ±0.25 (2) & 0.48 ±0.17 (2) & 0.44 ±0.26 (2) & 0.57 ±0.2 (2) & 0.13 ±0.04 (1) & 0.49 ±0.24 (2) \\
 & \cellcolor{red!25}NewsRoom (4) & 0.59 ±0.02 (4) & 0.59 ±0.03 (4) & 0.44 ±0.05 (4) & 0.55 ±0.03 (4) & 0.5 ±0.07 (4) & 0.36 ±0.06 (4) & 0.16 ±0.05 (4) & 0.45 ±0.04 (4) & 0.26 ±0.06 (4) & 0.21 ±0.08 (4) & -0.01 ±0.04 (0) \\
 & \cellcolor{red!25}SummEval (4) & 0.35 ±0.06 (4) & 0.44 ±0.14 (4) & 0.54 ±0.08 (4) & 0.38 ±0.02 (4) & 0.48 ±0.02 (4) & 0.19 ±0.06 (4) & 0.13 ±0.06 (4) & 0.29 ±0.09 (4) & 0.4 ±0.12 (4) & 0.15 ±0.05 (4) & 0.06 ±0.02 (2) \\
 & \cellcolor{red!25}WMT 2020 En-De (1) & 0.63 (1)  & 0.37 (1)  & 0.51 (1)  & 0.46 (1)  & 0.2 (1)  & 0.42 (1)  & 0.15 (1)  & 0.11 (1)  & 0.36 (1)  & 0.15 (1)  & -0.03 (1)  \\
 & \cellcolor{red!25}WMT 2020 Zh-En (1) & 0.54 (1)  & 0.39 (1)  & 0.48 (1)  & 0.41 (1)  & 0.25 (1)  & 0.42 (1)  & 0.15 (1)  & 0.14 (1)  & 0.39 (1)  & 0.15 (1)  & 0.01 (0)  \\
 & \cellcolor{red!25}WMT 2023 En-De (1) & 0.22 (1)  & 0.14 (1)  & 0.23 (1)  & 0.16 (1)  & 0.17 (1)  & 0.22 (1)  & 0.19 (1)  & 0.08 (1)  & 0.18 (1)  & -0.09 (1)  & -0.05 (1) \\
 & \cellcolor{red!25}WMT 2023 Zh-En (1) & 0.17 (1)  & 0.14 (1)  & 0.19 (1)  & 0.14 (1)  & 0.15 (1)  & 0.15 (1)  & 0.14 (1)  & 0.02 (1)  & 0.15 (1)  & 0.01 (0)  & 0.01 (0)  \\
\bottomrule
\end{tabular}}
\caption{Scores per dataset for all models we evaluate: Cohen’s kappa for categorical annotations and Spearman’s correlation for graded annotations. For Spearman's correlations, we report the number of significant correlations ($p<0.05$) for each model and dataset in brackets. Datasets in blue concern human-generated language while those in red concern model-generated text.}
\label{tab:all_models_results_table}
\end{sidewaystable*}

\begin{sidewaystable*}[ht!]\centering
\resizebox{0.9\textwidth}{!}{
\begin{tabular}{lllllllllllll}
\toprule
Type & Dataset (\#properties judged) & GPT-4o & Llama-3.1-70B & Gemini-1.5 & Mixtral-8x7B & Comm-R4 & Llama-3.1-8B & Mistral-7B & Starling-7B & OLMo-7B \\
\midrule
\multirow{17}{*}{\rotatebox[origin=c]{90}{Categorical Annotations}} & \cellcolor{blue!25}CoLa (1) & 0.35  & 0.41  & 0.45  & 0.47  & 0.3  & 0.35  & 0.51  & 0.39  & 0.26  \\
 & \cellcolor{blue!25}CoLa-grammar (63) & -0.04 ±0.06 & 0.35 ±0.25 & 0.33 ±0.23 & 0.21 ±0.16 & 0.05 ±0.09 & 0.24 ±0.21 & 0.19 ±0.19 & 0.16 ±0.16 & 0.04 ±0.06 \\
 & \cellcolor{blue!25}LLMBar-natural (1) & 0.86  & 0.86  & 0.71  & 0.62  & 0.37  & 0.55  & 0.56  & 0.46  & 0.21  \\
 & \cellcolor{blue!25}LLMBar-adversarial (1) & 0.67  & 0.92  & 0.32  & -0.07  & -0.25  & -0.3  & -0.29  & -0.25  & -0.05  \\
 & \cellcolor{blue!25}ToxicChat (2) & 0.42 ±0.1 & 0.37 ±0.03 & 0.41 ±0.36 & 0.33 ±0.21 & 0.33 ±0.26 & 0.22 ±0.03 & 0.41 ±0.07 & 0.33 ±0.16 & 0.31 ±0.2 \\
  & \cellcolor{red!25}Persona Chat (2) & 0.83 ±0.25 & 0.13 ±0.19 & 0.57 ±0.6 & 0.0 ±0.01 & 0.47 ±0.75 & -0.01 ±0.01 & -0.01 ±0.02 & -0.03 ±0.05 & -0.03 ±0.05 \\
 & \cellcolor{red!25}Topical Chat (2) & 0.57 ±0.61 & 0.09 ±0.13 & 0.03 ±0.04 & -0.02 ±0.03 & 0.48 ±0.74 & -0.0 & -0.03 ±0.05 & -0.03 ±0.05 & -0.03 ±0.05 \\
 & \cellcolor{red!25}ROSCOE-GSM8K (2) & 0.29 ±0.77 & 0.52 ±0.26 & 0.52 ±0.25 & -0.29 ±0.02 & -0.24 ±0.34 & 0.12 ±0.15 & 0.38 ±0.46 & -0.06 ±0.18 & -0.04 ±0.03 \\
 & \cellcolor{red!25}ROSCOE-eSNLI (2) & 0.05 ±0.16 & 0.1 ±0.09 & -0.01 ±0.01 & -0.03 ±0.05 & -0.04 ±0.04 & -0.04 ±0.04 & -0.01 ±0.09 & 0.06 ±0.17 & -0.03 ±0.06 \\
 & \cellcolor{red!25}ROSCOE-DROP (2) & -0.05 ±0.01 & 0.13 ±0.15 & -0.08 ±0.07 & -0.11 ±0.15 & -0.14 ±0.12 & -0.05 ±0.05 & -0.07 ±0.09 & -0.03 & -0.07 ±0.04 \\
& \cellcolor{red!25}ROSCOE-CosmosQA (2) & -0.29 ±0.06 & 0.03 & -0.26 ±0.09 & -0.29 ±0.12 & -0.3 ±0.05 & -0.11 ±0.16 & -0.25 ±0.2 & -0.09 ±0.16 & -0.23 ±0.12 \\
 & \cellcolor{red!25}QAGS (1) & 0.69  & 0.7  & 0.66  & 0.66  & 0.34  & 0.58  & 0.46  & 0.49  & 0.07  \\
 & \cellcolor{red!25}Medical-safety (2) & -0.01 ±0.09 & -0.02 ±0.08 & -0.01 ±0.09 & 0.03 ±0.07 & -0.0 ±0.01 & -0.02 ±0.01 & -0.02 ±0.07 & -0.01 ±0.01 & -0.01 ±0.09 \\
& \cellcolor{red!25}DICES-990 (1) & -0.22  & -0.15  & -0.16  & -0.14  & -0.1  & -0.16  & -0.08  & -0.08  & -0.12  \\
& \cellcolor{red!25}DICES-350-expert (1) & -0.3  & -0.26  & -0.06  & -0.02  & -0.13  & -0.07  & 0.06  & 0.01  & -0.04  \\
 & \cellcolor{red!25}DICES-350-crowdsourced (1) & -0.26  & -0.21  & -0.13  & -0.02  & -0.18  & -0.19  & 0.0  & -0.07  & 0.06  \\
& \cellcolor{red!25}Inferential strategies (1) & 0.47  & 0.4  & 0.25  & 0.13  & 0.01  & 0.04  & -0.01  & 0.12  & 0.02  \\
  \midrule
 \multirow{13}{*}{\rotatebox[origin=c]{90}{Graded Annotations}}
& \cellcolor{blue!25}Dailydialog (1) & 0.69 (1)  & 0.62 (1)  & 0.56 (1)  & 0.25 (1)  & 0.42 (1)  & 0.51 (1)  & 0.4 (1)  & 0.34 (1)  & 0.27 (1)  \\
 & \cellcolor{blue!25}Switchboard (1) & 0.6 (1)  & 0.48 (1)  & 0.53 (1)  & 0.07 (0)  & 0.38 (1)  & 0.17 (0)  & 0.36 (1)  & 0.35 (1)  & 0.08 (0) \\
 & \cellcolor{red!25}Persona Chat (4) & 0.2 ±0.09 (2) & 0.09 ±0.2 (1) & 0.11 ±0.06 (0) & 0.06 ±0.15 (0) & 0.17 ±0.21 (1) & -0.04 ±0.22 (1) & 0.04 ±0.13 (0) & -0.01 ±0.22 (1) & 0.06 ±0.18 (0) \\
 & \cellcolor{red!25}Topical Chat (4) & 0.22 ±0.02 (0) & 0.14 ±0.13 (1) & 0.11 ±0.1 (0) & 0.06 ±0.18 (1) & 0.1 ±0.14 (0) & 0.16 ±0.17 (1) & 0.25 ±0.05 (2) & 0.17 ±0.09 (1) & 0.08 ±0.13 (0) \\
 & \cellcolor{red!25}Recipe-generation (6) & 0.67 ±0.12 (6) & 0.64 ±0.14 (6) & 0.65 ±0.09 (6) & 0.42 ±0.18 (5) & 0.14 ±0.15 (2) & 0.31 ±0.15 (2) & 0.41 ±0.07 (6) & 0.34 ±0.2 (4) & 0.09 ±0.1 (0) \\
  & \cellcolor{red!25}ROSCOE-GSM8K (2) & 0.82 ±0.12 (2) & 0.81 ±0.12 (2) & 0.81 ±0.11 (2) & 0.81 ±0.13 (2) & 0.49 ±0.13 (2) & 0.8 ±0.11 (2) & 0.58 ±0.13 (2) & 0.64 ±0.15 (2) & 0.07 ±0.07 (0) \\
 & \cellcolor{red!25}ROSCOE-eSNLI (2) & 0.49 ±0.29 (2) & 0.39 ±0.38 (1) & 0.31 ±0.32 (1) & 0.31 ±0.09 (2) & 0.33 ±0.01 (2) & 0.23 ±0.17 (1) & 0.17 ±0.03 (1) & 0.19 ±0.13 (1) & -0.05 ±0.07 (0) \\
 & \cellcolor{red!25}ROSCOE-DROP (2) & 0.54 ±0.17 (2) & 0.55 ±0.19 (2) & 0.44 ±0.12 (2) & 0.29 ±0.14 (2) & 0.29 ±0.24 (1) & 0.44 ±0.07 (2) & 0.28 ±0.03 (2) & 0.17 ±0.15 (1) & -0.04 ±0.02 (0) \\
 & \cellcolor{red!25}ROSCOE-CosmosQA (2) & 0.57 ±0.21 (2) & 0.55 ±0.22 (2) & 0.56 ±0.11 (2) & 0.55 ±0.12 (2) & 0.62 ±0.15 (2) & 0.38 ±0.06 (2) & 0.54 ±0.21 (2) & 0.32 ±0.14 (2) & 0.3 ±0.14 (1) \\
 & \cellcolor{red!25}NewsRoom (4) & 0.57 ±0.05 (4) & 0.53 ±0.03 (4) & 0.53 ±0.05 (4) & 0.46 ±0.02 (4) & 0.19 ±0.06 (4) & 0.49 ±0.04 (4) & 0.19 ±0.04 (4) & 0.22 ±0.12 (3) & 0.09 ±0.06 (1)\\
 & \cellcolor{red!25}SummEval (4) & 0.45 ±0.11 (4) & 0.48 ±0.16 (4) & 0.33 ±0.03 (4) & 0.35 ±0.05 (4) & 0.17 ±0.05 (4) & 0.24 ±0.13 (4) & 0.38 ±0.1 (4) & 0.24 ±0.11 (4) & 0.03 ±0.07 (1) \\
  & \cellcolor{red!25}WMT 2020 En-De (1) & 0.57  & 0.44 (1)  & 0.38 (1)  & 0.39 (1)  & 0.13 (1)  & 0.34 (1)  & 0.33 (1)  & 0.3 (1)  & 0.0 (0)  \\
 & \cellcolor{red!25}WMT 2020 Zh-En (1) & 0.52 (1)  & 0.44 (1)  & 0.44 (1)  & 0.42 (1)  & 0.19 (1)  & 0.36 (1)  & 0.39 (1)  & 0.35 (1)  & 0.13 (1)  \\
 & \cellcolor{red!25}WMT 2023 En-De (1) & 0.2 (1) & 0.16 (1)  & 0.18 (1)  & 0.2 (1)  & 0.2 (1)  & 0.18 (1)  & 0.21 (1)  & 0.19 (1)  & -0.01 (0)  \\
 & \cellcolor{red!25}WMT 2023 Zh-En (1) & 0.15 (1)  & 0.13 (1) & 0.15 (1)  & 0.16 (1)  & 0.16 (1)  & 0.13 (1)  & 0.11 (1)  & 0.13 (1)  & 0.06 (0)  \\
\bottomrule
\end{tabular}
}
\caption{Scores per dataset for all models we evaluate using CoT prompts: Cohen’s kappa for categorical annotations and Spearman’s correlation for graded annotations. For Spearman's correlation, we report the number of significant correlations for each model and dataset in brackets. Datasets in blue concern human-generated language while those in red concern model-generated text.}
\label{tab:cot_models_results_table}
\end{sidewaystable*}

\end{document}